\documentclass{article} 
\usepackage{times} 
\usepackage{helvet} 
\usepackage{courier} 
\usepackage[hyphens]{url} 
\usepackage{graphicx} 
\usepackage{graphicx}  
\usepackage{microtype}
\usepackage{graphicx}
\usepackage{subfigure}
\usepackage{times}
\usepackage{soul}
\usepackage{url}
\usepackage[utf8]{inputenc}
\usepackage{graphicx}
\usepackage{amsmath,amssymb}
\usepackage{booktabs}
\usepackage{algorithm}
\usepackage{algorithmic}
\usepackage{shortcuts}
\usepackage{bbm}
\usepackage{xcolor}
\usepackage{array}
\usepackage{makecell}
\usepackage{fullpage}
\usepackage{natbib}
\usepackage{shortcuts}
\usepackage{authblk}
\graphicspath{{figures/}}

\newcommand{\yrcite}[1]{\citeyearpar{#1}}
\renewcommand{\cite}[1]{\citep{#1}}

\title{Rapidly Personalizing Mobile Health Treatment Policies\\ with Limited Data}
\author[1*]{Sabina Tomkins}
\author[2]{Peng Liao}
\author[1]{\\Predrag Klasnja}
\author[1]{Serena Yeung}
\author[2]{Susan Murphy}
\affil[1]{Stanford University}
\affil[2]{Harvard University}
\affil[2]{University of Michigan}
\affil[*]{\small{ Corresponding author: \texttt{stomkins@stanford.edu}}}
\date{} 
\begin{document}
\maketitle




\begin{abstract}
In mobile health (mHealth), 
reinforcement learning algorithms that adapt to one's context 
without learning personalized policies might fail to distinguish between the needs of individuals. 
Yet the high amount of noise due to the in situ delivery of mHealth interventions 
can cripple the ability of an algorithm to learn when  given access to only a single user's data, making  personalization challenging. 
 We present \ourapproach{}, which
 learns personalized policies via an adaptive, principled use of other users' data.  
We show that \ourapproach{} achieves an average of 26\% lower regret than  state-of-the-art  across all generative models. Additionally, we inspect the behavior of this approach 
in a live clinical trial, demonstrating its ability to learn from even a small group of users.

\end{abstract}
%!TEX root = pooling_icml.tex
\section{Introduction}
\label{sec:intro}

Mobile health (mHealth) interventions deliver  treatments to users
to support  healthy behaviors. These interventions offer an opportunity 
for social impact in a diverse range of domains
from substance abuse \cite{rabbi2017sara}, to disease management \cite{hamine2015impact} to physical inactivity \cite{consolvo2008activity}.
For example, to help users increase their physical activity, an mHealth application
might send a walking suggestions at times and in locations  when a user is likely to be able  
to pursue the suggestions. 
 The promise of mHealth hinges on the ability to provide interventions at times when users need the support \textit{and} are receptive to it \cite{nahum2017just}.
 Consequently, in developing reinforcement learning (RL) algorithms for mHealth our goal 
 is to be able to learn an optimal policy of when and how to intervene for a given user and context.

A significant challenge to learning an optimal policy  is that there are often  only a few opportunities per day to provide treatment. Furthermore,  wearable sensors  provide noisy estimates of critical metrics such as step counts  \cite{kaewkannate2016comparison}. 
In mHealth settings, it is critical for an algorithm to learn quickly,  in spite of noisy measurements and limited treatment data,
as   a poor policy can decrease user  engagement and potentially increase the risk of a user leaving a trial or otherwise abandoning treatment.  
Standard reinforcement learning algorithms can learn poorly in these settings.
Yet, demonstrations  of the effectiveness of these approaches (especially through live clinical trials) are lacking, 
and essential to establishing their feasibility.

We  present a  personalized RL algorithm developed to meet the challenges of  mHealth domains. Critically, we also evaluate its viability 
 with a live clinical trial.
To accelerate learning under the challenge of limited data we propose an approach that intelligently pools  data from all users, according to a  hierarchical Bayesian model,
so as to more quickly learn an optimal policy for each. 
We use empirical Bayes to update the model hyper-parameters.
This ensures that our approach is adaptive
in that for each user the extent to which their own data (relative to data from the entire population) informs their policy is updated over time.

 To inform the design of this clinical trial, we first conducted a smaller
 physical activity trial with sedentary individuals which we refer to as \TrialVone{}. 
In \TrialVone{}, contextual data was collected from each user's fitness tracker and smartphone. 
We use this data to construct a simulation environment to evaluate our 
 approach.
By mirroring  aspects of this trial 
we evaluate the algorithm in a challenging setting in which each user may experience the treatments a few times per day and in which the data is noisy.
As similar settings 
 exist beyond mHealth
and there is a dearth of acceptable methods to contend with their challenges, 
we propose our approach 
as a general framework for principled pooling in RL algorithms. 
Our main contributions are:
\begin{itemize}
\item[--] \ourapproach{}:  \textit{A Thompson Sampling algorithm for rapid personalization in limited data settings}. 
This algorithm employs empirical Bayes to adaptively adjust the degree to which policies are personalized to each user.
We present an analysis of this adaptivity in \secref{subsec:intuition} showing that \ourapproach{} learns to personalize to a user
as a function of the observed variance in the treatment effect both between and within users. 
\item[--] \textit{An empirical evaluation of our approach in a simulation environment constructed from  mHealth data}. \ourapproach{}
not only achieves 26\% lower regret than state-of-the-art, it also is better able to adapt to the degree of heterogeneity present in a population. 
\item[--] \textit{Evidence of the practicality of our approach from a live clinical trial}. A driving motivation of this work is to provide a reinforcement learning algorithm 
that can face the challenges of limited data in noisy online settings. We demonstrate that \ourapproach{} can be executed in a real-time online environment.  
\end{itemize}

%!TEX root = pooling_icml.tex
\section{Related Work}
\label{sec:related_work}

In mHealth several algorithms have been proposed for learning treatment policies.  These have typically followed two main paradigms. The first is learning a treatment policy for each user separately, such as \cite{rabbi2015mybehavior}, \cite{jaimes2016preventer}, and \cite{forman2018can}. This approach makes sense when users are highly heterogeneous, that is, their optimal policies differ greatly one from another.   However, this situation can present challenges for learning the policy when data is scarce and/or noisy, as in our motivating example of encouraging activity in an mHealth study where only a few decision time-points occur each day. The second paradigm is learning one treatment policy for all users
both in bandit algorithms  \cite{bouneffouf2012hybrid,paredes2014poptherapy,yom2017encouraging}, and in full reinforcement learning algorithms \cite{clarke2017mstress,zhou2018personalizing}. This second approach can potentially learn quickly but may result in poor outcomes if the optimal policies differ much between users.
In this work we demonstrate that a pooled approach has advantages over each of these paradigms. When users are heterogenous, our method achieves lower regret than batch approaches, and more quickly than personalized approaches. When users are homogenous our method performs as well as the batch approach.

Our proposed algorithm uses a mixed (random) effects Gaussian process (GP) model as part of a Thompson Sampling algorithm. While Gaussian process models have been used for multi-armed bandits \cite{chowdhury2017kernelized,brochu2010portfolio,srinivas2009gaussian,desautels2014parallelizing,wang2016optimization,djolonga2013high,bogunovic2016time} , and 
for contextual bandits  \cite{li2010contextual,krause2011contextual}, there is no work establishing their success in a setting with the challenges posed by mHealth. 
Furthermore, though our use of a mixed-effects GP resembles  that of \cite{shi2012mixed,luo2018mixed}  we consider a mixed-effects model in the context of RL rather than the previously considered prediction setting.

While we propose a bandit approach that pools across users in a structured manner, others have proposed pooling in other ways:
 Deshmukh et al. \yrcite{deshmukh2017multi} pool data from different arms of a single bandit, and Li and Kar \yrcite{li2015context} use context-sensitive clustering to produce aggregate reward estimates for the UCB bandit algorithm. 
 More relevant to this work are multi-task GPs, e.g. \cite{lawrence2004learning,bonilla2008multi,wang2012nonparametric}, however these have been proposed in the prediction as opposed to the RL setting. 
The Gang of Bandits \cite{cesa2013gang,vaswani2017horde} approach has been shown to be successful when there is prior knowledge on the similarities between users. For example, a social network graph might provide a mechanism for pooling. 
In contrast, our approach does not require prior knowledge of relationships between users and we adaptively update the degree of personalization. 
 
%!TEX root = pooling_icml.tex
\section{Our Approach}
\label{sec:approach}

We present an approach for learning personalized treatment policies in  mHealth settings,
 where a policy takes the user's  current context as input and outputs a treatment.
For example, context might include current location/weather while a treatment might be a physical activity message.
 Here, 
our goal is to learn such policies within a clinical trial in which users  enroll incrementally.
During the trial the developed algorithm  will learn a policy for each user based on the user's prior data as well as  data from  current and past users.  

\subsection{Problem setting}

\label{sec: notation}

Let $i \in [N] = \{1, \dots, N\}$ be the user index. For each user, we use $k \in \{1, 2, \dots\}$ to index decision times, i.e., times at which a treatment could be provided.   
Denote by $S_{i, k}$ the contextual features at the $k$-th decision time of user $i$, such as location.  
Let $A_{i, k}$ be the selected treatment. For simplicity, we consider  binary treatment $\activityset = \{0,1\}$. 
Recall that users enter the trial in staggered fashion. We denote by $t_{i, k}$ the calendar time of user $i$'s $k$-th decision time.

Our objective is to learn individual treatment policies for $N$ individuals; we treat this as $N$ contextual bandit problems.
We note that maintaining $N$ separate problems is important in settings such as ours where the true context is only sparsely observed and there is significant unobserved heterogeneity among different users. 
\secref{sec:bandit_formulation} reviews two approaches for using  Thompson Sampling \cite{agrawal2012analysis} and \secref{sec:pooling_method} presents our approach  for learning the treatment policy for any specific user. 

\subsection{Two Thompson Sampling instantiations}
\label{sec:bandit_formulation}

First consider learning the treatment policy separately per person. We refer to this approach as \none{}. 
At each decision time $k$, we would like to select a treatment  $\activity_{i, k} \in \{0,1\}$ based on the context $\state_{i, k}$.  
We model the reward $\reward_{i, k}$  by a Bayesian linear regression model: for user $i$ and time $k$
\begin{eqnarray}
{
R_{i, k} = \phi(S_{i, k}, A_{i, k})^\transpose \theta_{i} + \epsilon_{i, k}}
\label{bayes_reg}
\end{eqnarray}
where  $\phi(s, a)$ is a feature vector of context and treatment variables that are predictive of rewards (e.g. those described in \secref{sec:implementation}), 
$\theta_i$ is a parameter vector which we will learn, and $\epsilon_{i, k} \sim \mathbf{N}(0, \sigma_{\epsilon}^2)$ is the error term.  
The parameters $\{\theta_i\}$ are assumed independent across users and to follow a common prior distribution $\theta_i \sim \mathbf{N}(\mu_\theta, \Sigma_\theta)$.

Now at any decision time $k$, given the user's history so far  $\D_{i, k} = \{(S_{i, o},A_{i, o},R_{i, o}): o \leq k-1\}$ and the current context $S_{i, k}$, we use Thompson Sampling  to select the  treatment.  That is,  select treatment $\activity_{i, k} = 1$ with probability $\policy_{i, k}$: 
\begin{align}
\pi_{i, k}= \textrm{Pr} \{ \phi(S_{i, k}, 1)^\transpose \tilde \theta_{i, k} > \phi(S_{i, k}, 0)^\transpose \tilde \theta_{i, k}\} 
\end{align}
where $\tilde \theta_{i, k}$ follows the posterior distribution of $\theta_i$ given $\mathcal{D}_{i, k}$.  We note that the posterior distribution of $\theta_i$ is formed based on the user's own data.

In many  mHealth applications, the combination of noisy data and low numbers of decision point observations per day means that learning the treatment policy separately for each user can cause   slow policy improvement.  This motivates  leveraging data collected from other users to improve learning the optimal treatment policy for each user.  
A straightforward approach  is to learn a common bandit model for all users. In this setting, there are no individual-level parameters. The model is a single Bayesian regression model:
\begin{align}
\reward_{i, k}= \phi(\state_{i, k},\activity_{i, k})^T \theta + \epsilon_{i, k}. \label{compete model}
\end{align}
Note that $\theta$ does not vary by user. We then use the posterior distribution of the parameter $\theta$ to sample treatments for each user. 
This approach, which we refer to as \complete{}, may suffer from high bias when there is significant heterogeneity among users. This motivates our proposed method.

\subsection{Intelligent pooling across bandit problems}
\label{sec:pooling_method}

In our approach, which we call \ourapproach{},  we pool information across users in an adaptive way, i.e., when there is strong homogeneity  observed in the current  data, the method will pool more from others than when there is strong heterogeneity.

\subsubsection*{Bayesian random effects model}
Consider the Bayesian linear regression model (\ref{bayes_reg}). 
Instead of considering the $\theta_{i}$s as separate parameters to be estimated, we impose a random-effects  structure \cite{raudenbush2002hierarchical,laird1982random} on $\theta_{i}$:
\begin{eqnarray}
{
\theta_{i} = \theta_{pop} + u_i \label{random_effect} 
}
\label{randomeffect}
\end{eqnarray}
$\theta_{pop}$ is a population-level parameter and $u_i$
represents the person-specific deviation from $\theta_{pop}$ for user $i$.

We use the following prior for this model:
(1) $\theta_{pop}$ has prior mean $\mu_{\theta}$ and variance $\Sigma_{\theta}$, (2) $u_i$ has mean $\mathbf{0}$ and covariance $\Sigma_u$, 
and
 (3) $u_i \independent u_j$ for $i \neq j$ and $\theta_{pop} \independent \{u_i\} $ .

The variables $\mu_{\theta}, \Sigma_{\theta}$
 as well as the  variance of the person-specific effect $\Sigma_u$, 
and the residual variance $\sigma_\epsilon^2$ are hyper-parameters. In the prior  (Eqn. \ref{randomeffect}), we assume the person-specific effect on each element of the parameters $\theta$.
 In practice, one can use domain knowledge to specify which of the parameters should have the person-specific deviations; this will be the case in the experiments below.

We denote by $\mathcal{T}$ the set of times that the posterior distribution is updated. Specifically, let $T \in \mathcal{T}$ be an updating time and $\U_T \subseteq [N]$ be the set of users that are currently in or have finished the trial.  The history available at time $T$ is $\D_T = \{ (S_{i, k}, A_{i, k}, R_{i, k}, i): i \in \U_T, t_{i, k} \leq T \}$.   Suppose the number of tuples in  $\D_T$ is $n_T$.

The posterior distribution of each $\theta_{i}$ is Gaussian with mean and variance determined by a kernel function $K$ induced by the 
random effects 
model (Eqns.~\ref{bayes_reg}, \ref{random_effect}):
for any two tuples in $\D_T$, e.g., $x_l = (S^{(l)}, A^{(l)}, R^{(l)}, i_l), l = 1, 2$
\begin{align}
K_{}(x_1, x_2) & =\phi_1^\transpose  (\Sigma_{\theta} + \indicator{i_1 = i_2} \Sigma_{u} )  \phi_2
\label{our_kernel}
\end{align}
where $\phi_l = \phi(S^{(l)}, A^{(l)})$.
Note that the above kernel depends on $\Sigma_{\theta}$ and $\Sigma_u$.  The kernel matrix $\mathbf{K}_{n_T}$ is of size $n_T \times n_T$ and each element is the kernel value between two tuples in $\D_T$. The posterior mean and variance of $\theta_{i}$ given $\D_T$ can be calculated by
\begin{align}
\begin{split}
\mu_{i, T}  &= \mu_\theta + M_{i}^\transpose  (\mathbf{K}_{n_T} + \sigma_{\epsilon}^2 I_T)^{-1} \tilde R_{n_T}\\
\Sigma_{i, T}  &= \Sigma_{\theta} + \Sigma_u  - M_{i}^\transpose  (\mathbf{K}_{n_T} + \sigma_{\epsilon}^2 I_{n_T})^{-1} M_{i}
\end{split}
\label{post cal}
\end{align}
where 
$\tilde R_{n_T}$ is the vector of the rewards centered by the prior means, i.e., each element corresponds to a tuple $(S, A, R, j, h)$ in $\D_T$ given by $R - \phi(S, A)^\transpose \mu_{\theta}$, and 
 $M_{i}$ is a matrix of size $n_T$ by $p$, with each row corresponding to a tuple $(S, A, R, j)$ in $\D_T$ given by $\phi(\state, \activity)^\transpose ( \Sigma_{\theta} + \indicator{j = i} \Sigma_{u} )$.

\subsubsection*{Treatment selection}
To select a treatment for user $i$ at the $k$-th decision time, we use the posterior distribution of $\theta_{i}$  formed at the most recent update time $T$.  
That is, for the context $S_{i, k}$ of user $i$ at the $k$-th decision time,  \ourapproach{} selects the treatment $A_{i,k} = 1$  with probability
\begin{align}
\pi_{i, k}= \textrm{Pr} \{ \phi(S_{i, k}, 1) ^\transpose \tilde \theta_{i,T} > \phi(S_{i, k}, 0)^\transpose \tilde \theta_{i, T}\}  \label{prob}
\end{align}
where $\tilde \theta_{i,k} \sim \mathbf{N}(\mu_{i, T}, \Sigma_{i, T})$.

\subsubsection*{Updating hyper-parameters}
Thus far the degree of pooling across users has been determined by the choice of the hyper-parameters.   
While the prior mean $\mu_{\theta}$  and variance $\Sigma_{\theta}$ of the population parameter $\theta_{pop}$ can be set according to previous data or domain knowledge, it is difficult to pre-tune the variance components of the random effects.  Also the influence of the prior mean and variance on the Thompson Sampling algorithm decreases as data accrues and is used by the algorithm. 
However the influence of the variance components for the random effects on the degree of pooling persists even with increasing user data.  Thus at the update times, we use an empirical Bayes \cite{carlin2010bayes} approach to update $\lambda = (\Sigma_u, \sigma_\epsilon^2)$.  The updated values 
 maximize the marginal log-likelihood of the observed reward, marginalized over the population parameters $\theta_{pop}$ and the random effects. At every update time, $T$, we set the hyper-parameters as $\hat \lambda = \argmax l(\lambda | \D_T )$,  the maximizer of the marginal likelihood $l(\lambda | \D_T)$:
\begin{align}
\label{marginal likelihood}
\begin{split}
& l(\lambda | \D_T)  = -\frac{1}{2} [ \tilde R_{n_T}^\transpose (\mathbf{K}_{n_{T}}(\lambda) + \sigma_\epsilon^2 I_{n_T})^{-1} \tilde R_{n_T} \\
& \qquad  + \log \det (\mathbf{K}_{n_{T}}(\lambda)  + \sigma_\epsilon^2 I_{n_T}) + n_T \log(2\pi) ]
\end{split}
\end{align}
where  $\mathbf{K}_{n_{T}}(\lambda)$ is the kernel matrix as a function of parameters $\lambda = (\Sigma_{u}, \sigma_{\epsilon}^2)$. 
 \ourapproach{} is outlined in \algref{pooledalg}.

\begin{algorithm}
\caption{Thompson Sampling with Intelligent Pooling (\ourapproach{})\label{pooledalg}}
\begin{algorithmic}[1]
  \scriptsize
\STATE Initialize the posterior distribution $\{post(i)\}$ by the prior 
 \FOR {$ \ctimeindex \in [0, T_{}] $}
	\STATE Receive user index $i$ and  decision time index $k$
    \STATE Collect  state variable $\state$
    \STATE Obtain posterior distribution $post(i)$  of  $\theta_{i}$
    \STATE Calculate randomization probability $\pi$ in Eqn \ref{prob}
    \STATE Sample  treatment $\activity \sim  \operatorname{Bern} \left({\pi}\right)$
    \STATE Collect  reward $R$
    \STATE $\D \leftarrow \D \cup \{S, A, R, i\}$
 
    \IF{$\ctimeindex \in \mathcal{T}$}
    \STATE Update the hyper-parameters: $\lambda = \argmax l(\lambda | \D)$ in Eqn \ref{marginal likelihood}
   \STATE Update the posterior $post(\cdot) = post(\cdot| \D, \lambda)$ by Eqns \ref{post cal}
  \ENDIF 
 \ENDFOR
\end{algorithmic}
\end{algorithm}

\subsection{Impact of hyper-parameters }
\label{subsec:intuition}

Ideally, \ourapproach{} should learn to pool adaptively based on the users' heterogeneity. That is, the person-specific random effect should outweigh the population term if users are highly heterogenous. If users are highly homoegenous,  the person-specific random effect should be outweighed by the population term.  The amount of pooling is controlled by the hyper-parameters, e.g., the variance components of the random effects. 

To gain intuition, we consider a simple setting where the feature vector $\phi$ in the reward model (Eqn. \ref{bayes_reg}) is one-dimensional (i.e., $p =1$) and there are only two users (i.e., $i=1,2$).  Denote the prior distributions of population parameter $\theta_{pop}$ by $\mathbf{N}(0, \sigma_{\theta}^2)$ and the random effect $u_i$ by $\mathbf{N}(0,  \sigma_u^2)$.  Below we investigate how the hyper-parameters (e.g., $\sigma_u^2$ in this simple case), impact the posterior distribution.

Let $k_i$ be the index of decision time of user $i$ at the updating time $T$. 
In this simple setting, the posterior mean of $\theta_1$ can be calculated explicitly by 
	\[
\mu_{1} =   
	 \frac{[\delta \gamma + (1-\gamma^2) S_2] Y_1 + \delta \gamma^2 Y_2}{(1-\gamma^2)S_1 S_2  + \delta \gamma(S_1 + S_2) + (\delta \gamma)^2}
	\]
where  for $i=1,2$, $S_i = \sum_{k=1}^{k_i} \phi(A_{i, k}, S_{i, k})^2$,  $Y_i = \sum_{k=1}^{k_i}  \phi(A_{i, k}, S_{i, k}) R_{i, k}$, $\gamma =  \sigma_\theta^2/(\sigma_\theta^2 + \sigma_u^2)$ and $\delta = \sigma_\epsilon^2/\sigma_\theta^2$.	Similarly, the posterior mean of $\theta_2$ is given by
\[
\mu_{2} =
\frac{[\delta \gamma + (1-\gamma^2) S_1] Y_2 + \delta \gamma^2 Y_1}{(1-\gamma^2)S_1 S_2  + \delta \gamma(S_1 + S_2) + (\delta \gamma)^2}
\]

\begin{figure}
		\centering
		\includegraphics[width=.7\linewidth]{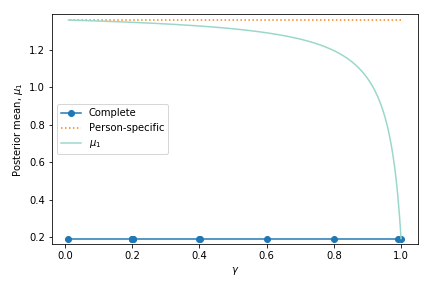}

		\caption{The posterior mean of $\theta_i$, $\mu_1$. As the variance of random effect $\sigma_u^2$ decreases, $\gamma$ increases and the posterior mean approaches the population-informed estimation (Complete) and departs from the person-specific estimation (Person-specific).  		\label{figexample} }
	\end{figure}

	When $\sigma_u^2 \goes 0$ (i.e., the variance of person-specific  effect goes to 0), we have $\gamma \goes 1$ and both posterior means $$\mu_{1}, \mu_{2} \goes \frac{Y_1 + Y_2}{S_1 + S_2 + \delta},$$
	which is the posterior mean under the model \complete ~(Eqn \ref{compete model}) using prior $\mathbf{N}(0,  \sigma_\theta^2)$. On the other hand, when $\sigma_u^2 \goes \infty$, we have $\gamma \goes 0$ and 
	$$
	\mu_1 \goes \frac{Y_1}{S_1}, ~\mu_2 \goes \frac{Y_2}{S_2}
	$$
	where correspond to the person-specific estimation of $\theta_1$ and $\theta_2$ under the model \none{} (Eqn \ref{bayes_reg}) using a non-informative prior. \figref{figexample} illustrates that when $\gamma$ goes from 0 to $1$, the posterior mean of $\theta_i$ smoothly transits from the population estimates to the person-specific estimates. 

%!TEX root = pooling_icml.tex
\section{Experiments}
\label{sec:experimental_design}
This work was conducted to prepare for deployment of  our algorithm in a live trial\footnote{For the purposes of anonymity we have redacted any identifying information about both trials mentioned here. We will provide these details upon acceptance.}.
Thus, to evaluate our approach we construct a simulation environment from a precursor trial,  \TrialVone{}. This simulation allows us to evaluate the proposed algorithm under various   settings that may arise in  implementation. 
 For example, heterogeneity in the observed rewards may be due to unknown subgroups across which users differ in their response to treatment. Alternatively,  this heterogeneity may vary across users in a more continuous manner.  We consider both scenarios in simulated trials. 
 In Sections \ref{sec:sim}-\ref{sec:empirical_eval} we evaluate the performance of \ourapproach{} against baselines and state-of-the-art. 
Having established its feasibility through a simulated environment,  in \secref{sec:clinical} we evaluate a pilot deployment of \ourapproach{} in a clinical setting.

 \subsection{Simulation environment}
 \label{sec:sim}
 %!TEX root = pooling_icml.tex
\TrialVone{} data is used to construct all features within the environment\footnote{We will release the code for this environment upon acceptance.}, and to guide choices such as how often to update the feature values.
 $S_{i,k}$ and $R_{i,k}$ denote the context features and reward of user $i$ at time $k$, respectively. The reward is the log step counts in the thirty minutes immediately following a decision time.
Selecting treatment one corresponds to sending an activity-suggestion message which requires several minutes of a user's time. Alternatively, selecting treatment  zero corresponds to sending a less burden-some message suggesting a very brief (30 second) activity. 

\begin{figure}[H]
\centering

\begin{minipage}[b]{.95\columnwidth}
    \centering\includegraphics[width=10.0cm]{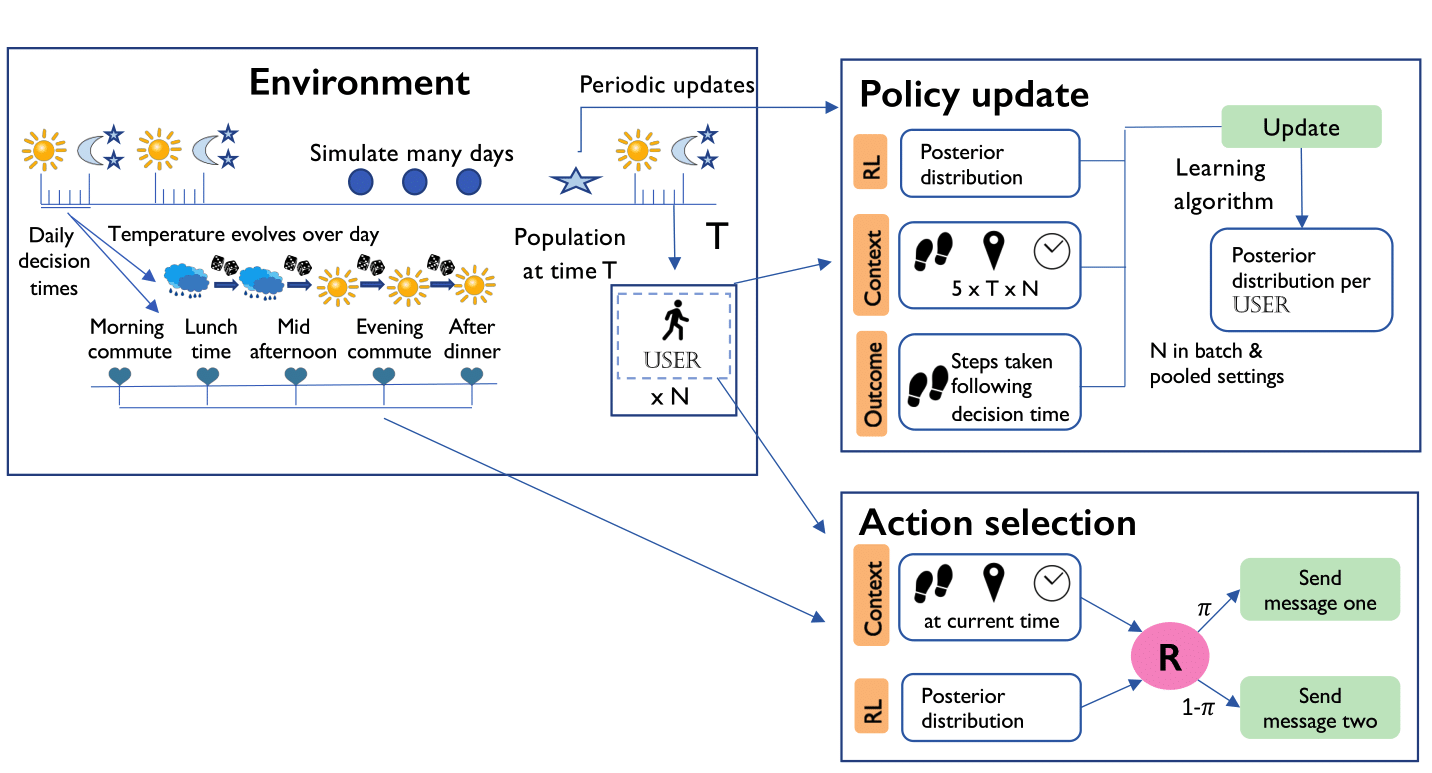}
    \caption{ Contextual features for a simulated \patient{} are composed of both general environmental features (such as time of day) and individual features (such as location). At
    decision times  a simulated user receives a message determined by the current treatment policy. Periodically this policy is updated according to a learning algorithm which outputs a new posterior distribution for each \patient. \label{sim_pic}}
    \end{minipage}
\end{figure}

\figref{sim_pic} describes the simulation while \tabref{user_states} describes  context features and rewards. 
Each context feature in \tabref{user_states} was constructed from  \TrialVone{} data.  
For example, we found that in \TrialVone{} data
splitting participants' preceding activity levels into the two categories of high or low best explained the reward. 

The temperature and location are updated throughout a simulated day according to probabilistic transition functions constructed from  \TrialVone{}.
The step counts for a simulated user are generated from  participants in \TrialVone{} as follows.   
We construct a one-hot encoding   containing the group-ID of a participant, the time of day, the day of the week, the temperature, the preceding activity level, and the location.
 Then for each possible realization of  the one-hot encoding we calculate the empirical mean  and empirical standard deviation of all step counts observed in  \TrialVone{}.
 
Let  $i$ denote the $i^{th}$ simulated user and $k$ denote a decision time.  This simulated user's context is encoded via the same one-hot encoding to produce 
 $h(\state_{i,k})$.  The corresponding  empirical mean  and empirical standard deviation from \TrialVone{} form
 $\mu_{h(\state_{i,k})}$  $\sigma_{h(\state_{i,k})}$ respectively.  
At non-decision times step counts are generated according to 
\begin{equation}R_{i,k} = \mathbf{N}(\mu_{h(\state_{i,k})},\sigma^2_{h(\state_{i,k})}).\label{not_avail}\end{equation}

\begin{table}[H]
\small
\centering
\resizebox{1.0\columnwidth}{!}{%
\begin{tabular}{|p{4.0cm}p{4.0cm}p{1.5cm}|}
\hline
    \multicolumn{3}{|c|}{ 
\normalsize{ \textbf{State Features}}} \\
\hline
    \multicolumn{1}{|l|}{\textbf{Name}}  &  \multicolumn{1}{l|}{\textbf{Value} }&  \multicolumn{1}{l|}{\textbf{ \makecell[l]{ \textbf{\patient}\\ Specific}  }}\\
\hline
Time of day & \makecell[l]{Morning(0) 9:00 and 15:00  \\ Afternoon(1)  15:00 and 21:00 \\ Night(2) 21:00 and 9:00} & No   \\
\hline
Day of the week & Weekday(0) or Weekend(1)  & No \\
\hline
Temperature& Cold(0) or Hot(1)&No \\
\hline
  Preceding activity level& Low(0) or High(1)& Yes \\
Location & Other(0) or Home/work(1)& Yes \\
Intercept & 1& Yes \\
\hline
    \multicolumn{3}{|c|}{
\normalsize{\textbf{Reward}} } \\
    \hline
Step count & Continuous on log scale& Yes \\
\hline
\end{tabular}
}
\caption{\textmd{The value used in encoding each feature is shown in parentheses. For example cold(0) indicates that cold is coded as a 0 wherever this feature is used. }
\label{user_states}}
\end{table}

\textbf{\populationgen{}} 
This model, which we denote \populationgen{}, allows us to compare the performance of the approaches under different levels of population heterogeneity. 
The step count after a decision time is a modification of \eqnref{not_avail} to reflect the interaction between  context and treatment on the reward and heterogeniety in treatment effect.
Let  $f(S_{i,k})\subseteq h(S_{i,k})$. 
Let $\beta$ be a vector of coefficients of $f(S_{i,k})$ which weigh the relative contributions of the entries of $f(S_{i,k})$ that interact with treatment on the reward. 
The magnitude of the entries of $\beta$  are set using \TrialVone{}. Step counts ($R_{i,k}$) are generated  as 
  \begin{equation}R_{i,k} = \mathbf{N}(\mu_{h(\state_{i,k})},\sigma^2_{h(\state_{i,k})})+ A_{i,k}( f(\state_{i,k})^T\beta_{i} + Z_i). \label{avail} 
  \end{equation}

The inclusion of $Z_i$ will allow us to evaluate the relative performance of each approach under different levels of population heterogeneity. 
Let $\beta^l_i$ be the coefficient of the location term for the $i^{th}$ user. 
We consider three scenarios (shown in \tabref{table:Z}) to generate $Z_i$, the person-specific effect, and $\beta^l_i$ the location-dependent person-specific effect. 
The performance of each algorithm under each scenario will be analyzed in \secref{sec:empirical_eval}.  In the smooth scenario, $\sigma$ is equal to the standard deviation of the observed treatment effects $[f(S_{i,k})^\transpose \beta\ :\ S_{i,k} \in \TrialVone{}]$ and $\beta^l_i$ is set to 0.1.

In the bi-modal scenario each simulated user is  assigned a base-activity level: low-activity users or high-activity users (these two groups were constructed from analyses of  \TrialVone{} using non-parametric clustering).  When a simulated user joins the trial they are placed into either group one or two with equal probability. The values of  $z_1,\beta^l_1$ and $z_2,\beta^l_2$ are set so that for all users in group 1, it is optimal to send a treatment ~75\% of the time while for all users in group 2 it is optimal to send a treatment ~25\% of the time. Group membership is not known to any of the algorithms. 

\begin{table}[H]
\centering	
\resizebox{\columnwidth}{!}{%
\begin{tabular}{|p{1.5cm |}|p{1.5cm |}|p{1.5cm |}||}
\hline
   \multicolumn{1}{|c|}{Homogeneous}&\multicolumn{1}{c|} {Bi-modal } & \multicolumn{1}{c|}{Smooth} \\
\hline
    \multicolumn{1}{|c|}{$Z^i =0$ $\beta^l_i$=0} &   \multicolumn{1}{r|}{ $ Z_i,\beta^l_i = \begin{cases}
      z_1, \beta^l_1 & \text{if}\ i \in \text{group one} \\
      z_2, \beta^l_2&  \text{if}\ i \in  \text{group two}
    \end{cases}$ } &  \multicolumn{1}{r|}{ $Z_i \sim \mathcal{N}(0,\sigma^2)$ $\beta^l_i\sim \mathcal{N}(0,\sigma_l^2)$} \\
    \hline
\end{tabular}}
\caption{\textmd{Settings for Z in three cases of homogeneous, bimodal and smoothly varying populations. } \label{table:Z}}

\end{table}

When a simulated user is at a decision time the user will receive a treatment according to whichever RL policy is being run through the simulation.

 \subsection{Simulation implementation details}
 \label{sec:implementation}
  %!TEX root = pooling_icml.tex

In \secref{sec:approach} we introduced the feature vector  $\phi$, recall that $\phi$ is the vector $\phi(\state_{i,k},\activity_{i,k}) \in \R^p$ used in  the model for the reward.  
The features in the reward model for all algorithms considered here are,
\begin{equation} \label{phi_feature_vector}
\begin{split}
\phi(\state_{i,k},A_{i,k})^T  = & \big( g(\state_{i,k},A_{i,k})^T, \pi_{i,k} f(\state_{i,k})^T, \\
&\qquad (A_{i,k}-\pi_{i,k})f(\state_{i,k})^T \big)
\end{split}
\end{equation}
where $g(S_{i,k})$  is a subset of $h(S_{i,k})$, containing:  an intercept term (equal to $1$), time of day, \text{day of the week}, \text{preceding activity level}, and \text{location} and $f(\state_{i,k})=g(S_{i,k})$.  Recall that the bandit algorithms produce $\pi_{i,k}$ which is the probability that $A_{i,k}=1$.

The inclusion of the term $\small{(A_{i,k}-\pi_{i,k})f(\state_{i,k})}$ is motivated by \cite{liao2016sample,boruvka2018assessing,greenewald2017action}, who demonstrated that 
action-centering can protect against mis-specification in the baseline effect (e.g., the expected reward under the action 0).   
 In $\TrialVone$ we observed that  users varied in their overall responsivity and that a user's location was related to their responsivity. In the simulation, we assume the person-specific random effect  on four parameters in the reward model (i.e., the coefficients of terms in $g$ and $f$ involving the intercept and location).

 Finally, we constrain the randomization probability to be within [0.1, 0.8] to ensure continual learning. The update time for the hyper-parameters is set to be every 7 days.
All approaches are implemented in Python and we implement GP regression with the software package GPytorch \cite{gardner2018gpytorch}.

  \subsection{Simulation results}
 \label{sec:empirical_eval}
  %!TEX root = pooling_icml.tex
 In this section, we present an empirical analysis of our algorithm ($\ourapproach$), comparing to two standard methods  $\complete$ and $\none$, which are outlined in \secref{sec:bandit_formulation}.  Recall that  \ourapproach{} includes person-specific random effects, as described in \eqnref{random_effect}. 
In $\none$,  all users are assumed to be different and there is no pooling of data and in 
 $\complete$, we treat all users the same and learn one set of parameters across the entire population.

Additionally, to assess \ourapproach's ability to pool across users we compare our approach to Gang of Bandits  \cite{cesa2013gang}, which we refer to as \gob. 
As this model requires a relational graph between users, we  construct a graph using the generative model \populationgen{} which connects users according to each of the three settings: homogenous, bi-modal and smooth.
For example, with knowledge of the generative model users can be connected to other users as a function of their $Z_i$ terms. 
As we will not have true access to the underlying generative model in a real-life setting we distort the true graph to reflect this incomplete knowledge.
That is we add ties to dissimilar users at 50\% of the strength of the ties between similar users.

Let $a^*_{i,k} = \indicator{f(S_{i,k})^T\beta_{i}^ *+Z_i \geq 0}$ be the optimal action for user $i$ at time $k$. We calculate the regret as
\begin{equation}
\text{regret}_{i,k}=|f(S_{i,k})^T\beta_{i}^ *+Z_i|  \indicator{a^*_{i,k}\neq A_{i,k}}  \label{eqn:regret}
\end{equation}
where $\beta^*_i$ is the optimal $\beta$ for the $i^{th}$ user.

In these simulations each trial has  32 users. 
Each user remains in the trial for 10 weeks and the entire length of the trial is 15 weeks, where the last cohort joins in week six.  The number of users who join each week is a function of the recruitment rate observed in $\TrialVone$. In all settings we run 50 simulated trials. 

\begin{figure}[H]
\centering

\begin{minipage}[b]{0.95\linewidth}

    \centering\includegraphics[width=9.0cm]{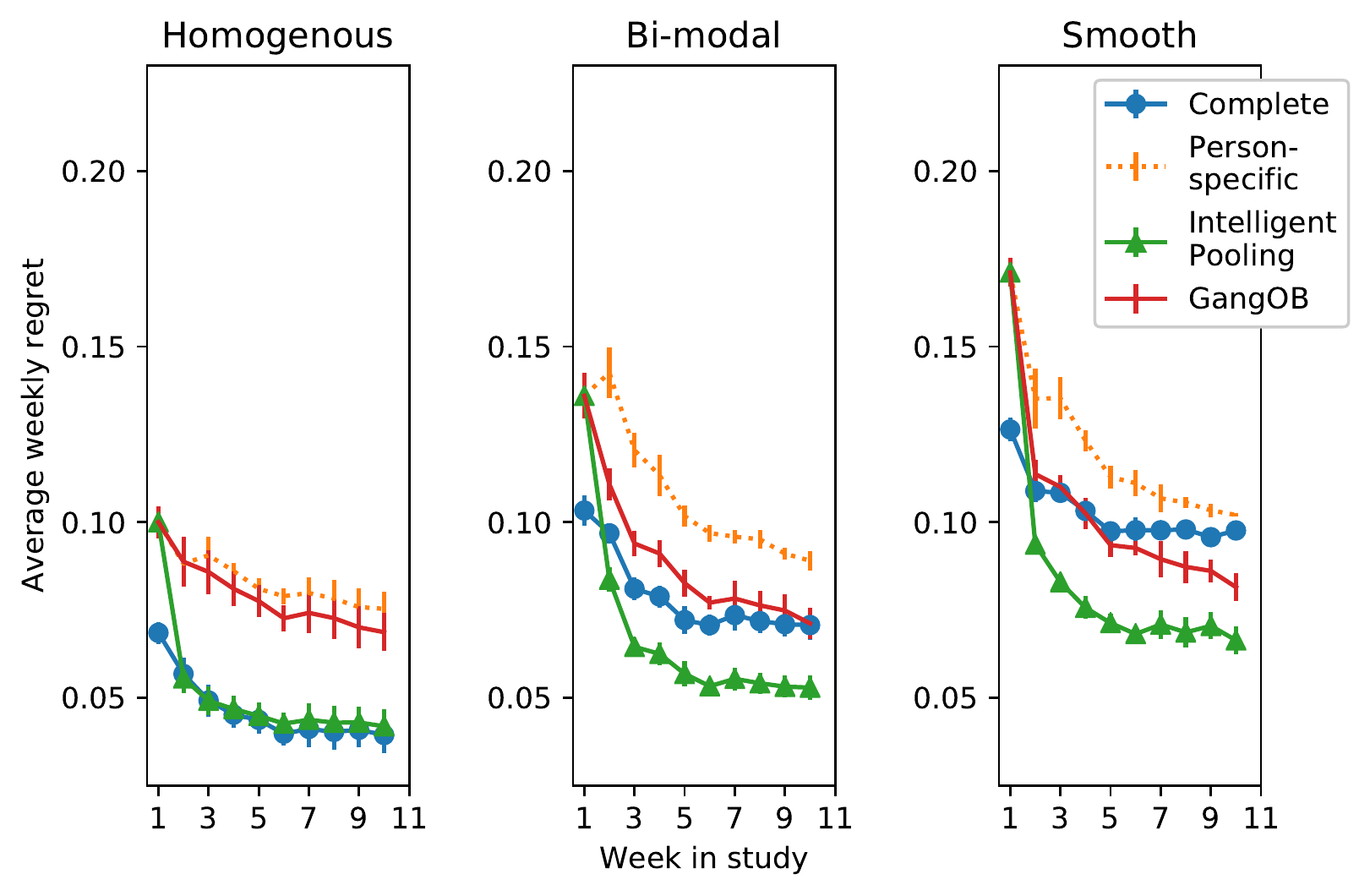}
    \caption{
\textbf{\populationgen{} generative model} 
 Regret averaged across all users for each week in the trial, i.e. average regret of all users in their first week of the trial.}\label{regret}
    \end{minipage}
  \end{figure}

First,  \figref{regret} provides the regret  averaged across all users across 50 simulated trials where the reward distribution follows the generative model $\populationgen$. 
Though users join the trial in a staggered fashion, so that in the first week of the trial only a few users are active,  the horizontal axis in \figref{regret} is the average regret over all users in their nth week in the trial, e.g. in their first week, their second week, etc. In the bi-modal setting there are two groups, where all users in group one have a positive response to treatment on average, while the users in group two have a negative response to treatment. An optimal policy would learn to not send interventions to users in the first group, and to send them to users in the second. To evaluate each algorithm's ability to learn this distinction we show the percentage of time each group received a message in \tabref{table:bi-modal}.

\begin{table}[H]
\centering
\resizebox{.65\columnwidth}{!}{%
\begin{tabular}{|p{1.75cm}|p{2.0cm }|p{2.0cm }|}
\hline
 & \makecell[l]{\small{Group one} \\ \small{optimal policy} \\\small{= send}}& \makecell[l]{\small{Group two} \\ \small{optimal policy} \\\small{= don't send}}  \\
\hline
\small{\complete} &    \multicolumn{1}{r|}{0.49} &   \multicolumn{1}{r|}{0.46}  \\
\hline
\makecell[l]{\small{\textsc{Person-}}\\ \small{\textsc{Specific}} }& \multicolumn{1}{r|}{0.65} &   \multicolumn{1}{r|}{0.49}\\
 \hline
  \makecell[l]{\small{\textsc{GangOB}} }& \multicolumn{1}{r|}{0.57} &   \multicolumn{1}{r|}{0.35}\\
   \hline
\makecell[l]{\small{\textsc{Intelligent-}}\\ \small{\textsc{Pooling} }}& \multicolumn{1}{r|}{0.59} &   \multicolumn{1}{r|}{0.36}     \\
 \hline
\end{tabular}
}
\caption{ Average fraction of times treatment was sent (action=1), over 50 simulations (bi-modal generative model $Z^b$).
}\label{table:bi-modal}
\end{table}

The relative performance of the approaches depends on the heterogeneity of the population. When the population is
very homogenous \complete{} excels, while its performance suffers as heterogeneity increases. \none{} is able to personalize; as shown by \tabref{table:bi-modal},
it can differentiate between individuals. However, it  learns slowly and can only approach the performance of \complete{} in the smooth setting of \populationgen{}
where users differ the most in their response to treatment. Both  \ourapproach{} and  \gobnoisy{} are more adaptive than either \complete{} or \none{}. \gobnoisy{} consistently outperforms \none{}
and achieves lower regret than \complete{} in some settings. In the homeogenous setting we see that \gob{} can utilize social information more effectively than \none{} does while in the smooth setting it can adapt to individual differences more effectively than \complete{}. Yet, \ourapproach{} demonstrates stronger and swifter adaptability than does \gob{}, consistently achieving lower regret at quicker rates. Finally, the algorithms differ in their suitability for real-world applications, especially when data is limited. 
 \gobnoisy{} requires  reliable values for hyper-parameters and can depend on  fixed knowledge about relationships between users.  
 \ourapproach{}   can learn how  to pool between individuals over time and without prior knowledge.

    \section{\ourapproach{}  Pilot}
     \label{sec:clinical}
  %!TEX root = pooling_icml.tex
The simulated experiments provide insights into the potential of this approach for a live deployment. As we see reasonable performance in the simulated setting, we now discuss an initial pilot deployment of \ourapproach{} in a physical activity clinical trial setting, which we refer to as \clinical. In   \clinical{},  following an initial  ten users in the  clinical trial \ourapproach{} is deployed  for each of the subsequent ten users.   At each decision time for these subsequent ten users, \algref{pooledalg} uses all data up to that decision time (i.e. from the initial ten users as well as from the subsequent ten users).

\begin{figure}[h!]
\centering
\begin{minipage}[b]{.95\columnwidth}
    \centering\includegraphics[width=10.0cm]{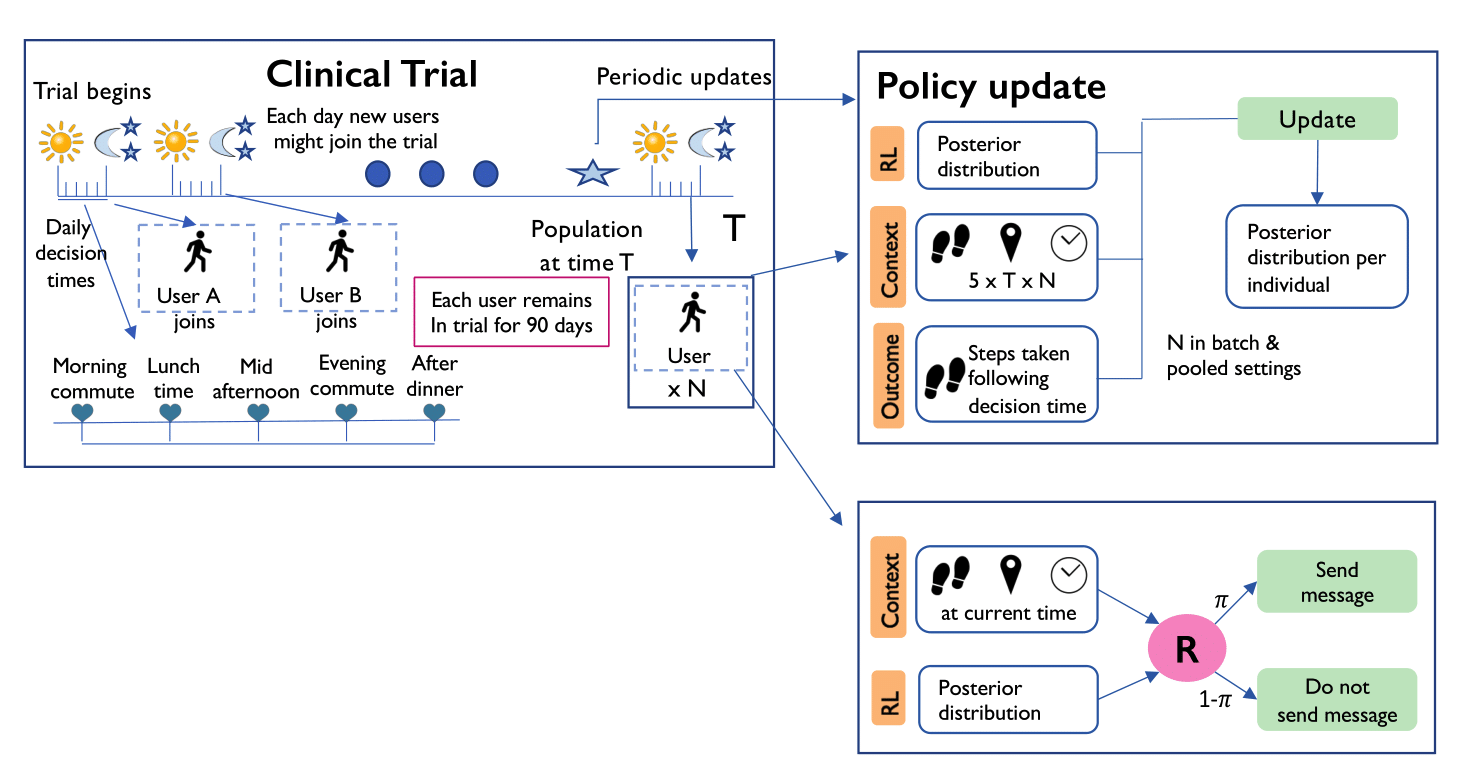}
    \caption{Setup of \clinical{}. Here we see that users can receive treatments up to five times a day and that each user remains in the trial for 90 days. Users enter the trial asynchronously. \label{clinical_pic}}
    \end{minipage}
\end{figure}

\figref{clinical_pic} provides details of this pilot study.
We use the Bayesian Thompson Sampling model shown in \secref{sec:implementation}. The features used in 
the trial are shown in in \tabref{clinical_states}. 
The feature \textit{engagement} represents the extent to which a user engages with the mHealth  application measured
as a function of how many screen views are made within the application. 
The feature \textit{dosage} represents the extent to which a user has received interventions.  This feature is designed to increase with 
exposure to intervention but can decline when a treatment is skipped. Thus, if a user did not receive treatment for a sufficient period of time
their dosage could be low. 
We provide a full description of these features in Section 3 of the supplement. 
As \clinical{} only includes a small number of users,   we deploy a simple model with two person-specific random effects on the intercept term  in $g$ and $f$ (\eqnref{phi_feature_vector}).

\begin{table}[H]
\small
\centering
\resizebox{1.0\columnwidth}{!}{%
\begin{tabular}{|p{4.0cm}p{4.0cm}p{1.5cm}p{1.5cm}|}
\hline
    \multicolumn{4}{|c|}{ 
\normalsize{ \textbf{State Features}}} \\
\hline
    \multicolumn{1}{|l|}{\textbf{Name}}  &  \multicolumn{1}{l|}{\textbf{Value} }&  \multicolumn{1}{l|}{\textbf{ \makecell[l]{ \textbf{\patient}\\ Specific}  }}&  \multicolumn{1}{l|}{\textbf{Included in $f$} }\\
\hline
\hline
Temperature& Continuous & Yes (based on location) & No \\
\hline
 Preceding activity level& Continuous& Yes  & No \\
 Variation in preceding activity level& Continuous& Yes & Yes \\
 Engagement with mobile application & Continuous& Yes & Yes  \\
Dosage & Continuous& Yes & Yes  \\
Location & Other(0) or Home/work(1)& Yes  & Yes \\
Intercept & 1& Yes&Yes \\
\hline
    \multicolumn{4}{|c|}{
\normalsize{\textbf{Reward}} } \\
    \hline
Step count & Continuous on log scale& Yes & NA\\
\hline
\end{tabular}
}
\caption{\textmd{State feature descriptions for \clinical.}
\label{clinical_states}}
\end{table}

\begin{figure}[h!]
\centering

\begin{minipage}{0.95\linewidth}
    \centering\includegraphics[width=8.0cm]{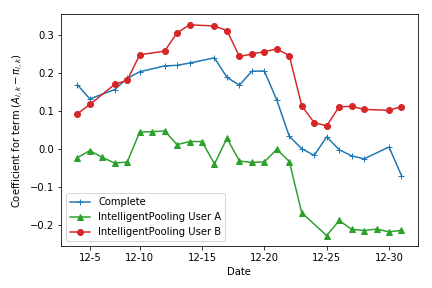}
    \caption{
    Posterior mean of the coefficient of $(A_{i,k}-\pi_{i,k})$ in \eqnref{phi_feature_vector}
for users A and B.}\label{two_users}
    \end{minipage}
  \end{figure}

\textbf{Personalization in \clinical}
By comparing how the decisions to treat under \ourapproach{} differ from those under \complete{}, 
 we provide preliminary
 evidence that \ourapproach{} personalizes to users.
\figref{two_users} displays the posterior mean of  the coefficient of the $A_{i,k}-\pi_{i,k}$ term in   $f$.  This coefficient represents 
the overall effect of treatment on user $i$.
During the prior 7 days the user has not experienced much variation in activity at this time and the user's engagement is low. 
Note that the treatment appears to have a positive effect on User B in this context whereas on User A there is little evidence of a positive effect.   If \complete{} had been used to determine treatment, user A  might have been over-treated.

\begin{figure}[h!]
\centering
\begin{minipage}{0.99\linewidth}
    \centering\includegraphics[width=8.0cm]{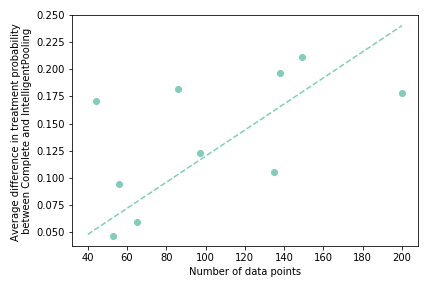}
    \caption{
The difference in treatment probabilities between \ourapproach{} and  \complete{} as a function of the amount of data from a user. Each dot is a different user. }\label{data_diff}
    \end{minipage}
  \end{figure}

 For each user we calculated the difference in treatment probabilities  between \ourapproach{} and  \complete{}. We see a weak linear trend with time (\figref{data_diff}), that is, as more data accumulates the difference between treatment probabilities under \complete{} and \ourapproach{} grows, with a Pearson correlation  coefficient of .56 ($p<.1$).  This is a signal that personalization strengthens as a user provides more  data. 

\begin{figure}[h!]
\centering

\begin{minipage}{0.99\linewidth}
    \centering\includegraphics[width=8.0cm]{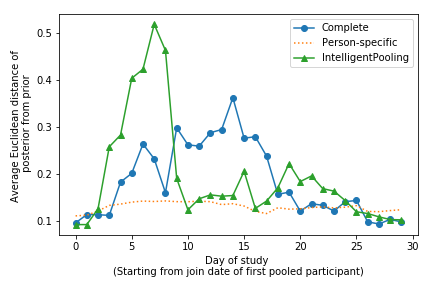}
    \caption{
Mean squared distance between posterior and  prior mean of the  coefficients of $(A_{i,k}-\pi_{i,k})f(S_{i,k})$.
} \label{speed}
    \end{minipage}
  \end{figure}

\textbf{Speed of policy learning in \clinical}
We consider the speed at which \ourapproach{} diverges from the prior, relative to the speed of divergence for \none.   \figref{speed} provides the Euclidean distance between the learned posterior and prior parameter vectors (averaged across the data from the 10 users at each time). 
From \figref{speed} we see that  
 \none{} hardly varies over time in contrast to \ourapproach{} and \complete{}, which suggests that \none{} learns more slowly.

%!TEX root = pooling_icml.tex
\section{Conclusion}
When data on individuals is limited a natural tension exists between personalizing (a choice which can introduce  variance)
and pooling (a choice which can introduce bias).
In this work we have introduced a novel algorithm for personalized
reinforcement learning, \ourapproach{} that presents a principled mechanism for balancing this tension.
 We demonstrate the practicality of our approach in the setting of mHealth.
 In simulation we achieve improvements of 26\% over a state-of-the-art-method, while in a live clinical trial 
we show that our approach shows promise of personalization on even a limited number of users.  
We view adaptive pooling as a first step in addressing the trade-offs between personalization and pooling.
 The question of how to quantify the benefits/risks for individual users is an open direction for future work.

\bibliography{pooling}
\bibliographystyle{myapalike}

\end{document}

\twocolumn[
\icmltitle{
Supplementary materials.
}
\frenchspacing
]

\section{Simulation}
We include additional information about the simulation environment. We would like to reiterate that if accepted we will publish the code for the entire environment to promote reproducibility. We first explain general information about the simulation environment. We then provide the procedures for generating state variables (features) in the simulation. Finally, we discuss how we used \TrialVone{} to arrive at the feature representations used in the simulation.

\textbf{Simulation dynamics} Within the simulation states are updated every thirty minutes. Each thirty minutes is associated with a date-time, thus we can acquire the month from the current time which is useful in updating the temperature.  The decision times are set roughly two hours apart from 9:00 to 19:00.

\textbf{Availability} In the real-study users are not always available to receive treatment for a suite of reasons. For example, they may be driving a vehicle or they might have recently received treatment. Thus, at each decision time we update the context feature $Available_i \sim Bernoulli(.8)$. for the $i^{th}$ user where $Available_i$ is drawn from a Bernoulli.  This condition reduces the distance between the settings in the environment and those in a real-world study. At each decision time interventions are only sent to users who are available; i.e.  user $i$ cannot receive an intervention when $Available_i =0$. 

\textbf{Recruitment} We follow the recruitment rate observed in $\TrialVone$. For example, if 20\% of the total number of participants were recruited in the third week of $\TrialVone$ we recruit 20\% of the total number of participants who will be recruited in the third week of the simulation. To explore the effect of running the study for varying lengths we scale the recruitment rates. For example, if the true study ran for 8 weeks, and we want to run a simulation for three weeks, we proportionally scale the recruitment in each of the three weeks so that the relative recruitment in each week remains the same.  
In these experiments we would like to recruit the entire population within 6 weeks. Thus about 10\% of participants are recruited each week, except for the second week of the study where about 30\% of all participants are recruited. This reflects the recruitment rates seen in the study, which were more of less consistent throughout besides one increase in the second week.

We generate states from historical data. Given relevant context we search historical data for states which match this given context. This subset of matching states can be used to generate new states. We discuss this in more detail in \secref{query}. Then, we describe in more detail how we generate temperature, location and step counts. 

\subsection{Querying history}
\label{query}
 \algref{alg:statefunctions} is used to obtain relevant historical data in order to form a probability distribution over some target feature value. 
 For example, if we would like a probability distribution over discretized temperature IDs under a given context, we would search over the historical data for all temperature IDs present under this context. This set of context-specific temperature IDs can then be used to form a distribution to simulate a new ID. This process  of querying historical data is used throughout the simulation and is outlined in \algref{alg:statefunctions}. For example, it is used in generating new step counts, new locations and new temperatures.

 \begin{algorithm}[H]
 \caption{$\statefunctions$}
 \label{alg:statefunctions}
 \begin{algorithmic}[1]
\STATE{\textsc{INPUT} =  historical data $   [\mathbf{x_i} ; i = [1,N] ]$, conditioning state $\mathbf{x^*}$, target data variable $y=f(x)$ }, 
\STATE{$\mathcal{S}= \{\}$}
\FOR{$i=1$ to $N$}
\IF{$\mathbf{x_i}==\mathbf{x^*}$}
\STATE{Add $f(x_i)$ to $\mathcal{S}$}
\ENDIF
\ENDFOR
\STATE{\textsc{OUTPUT} =$\mathcal{S}$}
\end{algorithmic}
 \end{algorithm}

As the simulation environment simulates draws stochastically from a variety of probability distributions, 
it is possible it draws a state which was not present in the historical dataset. In this case there is a process
for finding a matching state. Similarly we might have a state in the historical dataset with insufficient samples 
to form an informative (not overly-noisy) distribution. In this case we also find a surrogate state with which 
to generate future step counts.
The idea of the process is to find the closest state to the current state, such 
that this close state has sufficient data to generate a good distribution. 
Again, given a state, we want to be able to generate a step count from a distribution 
with sufficient data to inform its parameters. The pseudocode for how we do so is shown in \algref{findmatch}

This algorithm takes as input a target state, $s^*$. We also have a dictionary(hasmap) formed from the 
historical dataset. The keys to this dictionary are the states which existed in the dataset. A value is
an array of step counts for this state. 

\begin{algorithm}[H]
	\caption{\textsc{\small{FindMatch}}\label{findmatch}}
	\begin{algorithmic}[1]
		
		\STATE{\textsc{INPUT} =  current state $s^* \in \mathbb{R}^d$, dictionary of existing states to step counts $\mathbb{D}=\{s:[c_1,\dots,c_N]\}$} 
		\STATE{match$\leftarrow$None}
		\IF{$s^* \in \mathbb{D}$ and $len( \mathbb{D}[s^*])>30$}
		\STATE{match$\leftarrow s^*$}
		\ELSE
		\STATE{$new\_size$ = d-1}
		\WHILE{match is None}
		\STATE{\#find state of size new size with most data points in historical dataset}
		\STATE{form new states of size $new\_size$}
		\STATE{rank states $s$ by $len( \mathbb{D}[s])$}
		\STATE{choose state with greatest len}
		\STATE{$temp\leftarrow max_s len( \mathbb{D}[s])$}
		\IF{$ \mathbb{D}[temp] > 30$}
		\STATE{match$\leftarrow temp$}
		\ENDIF
		\STATE{$new\_size=new\_size-1$}
		\ENDWHILE
		\ENDIF
	\end{algorithmic}
\end{algorithm}

This procedure gives the closest state with the most data points to our current state.

To be more explicit about lines 8-11. A state is a vector of some length, for example $[1,0,1]$. 
When we consider all subsets of size 2, we are considering the subsets $[1,0]$,$[1,1]$, and $[0,1]$. 
For each of these we can look in the historical data set and find all points where this state was true. 
Thus for each subset we'll get a new list of points,  $[1,0] = [c_1,\dots,c_{N1}]$ $[1,1] = [c_1,\dots,c_{N2}]$,
$[0,1] = [c_1,\dots,c_{N3}]$. We now look at $N1,N2,N3$ and choose the state with the highest value. For example, 
if the lists were:  $[1,0] = [c_1,\dots,c_{100}]$ $[1,1] = [c_1,\dots,c_{2}]$,
$[0,1] = [c_1,\dots,c_{300}]$, we would choose $s=[0,1]$.
Now if we encounter the state $[1,0,1]$ and there is insufficient data to form a distribution from this state, 
we will instead form it from the values found under the state $[0,1]$, $[c_1,\dots,c_{300}]$.

\subsection{Generating temperature}
We mimic a trial where 
everyone resides in the same general area, such as a city. In this setting everyone experiences the same global temperature. 
We describe how to obtain temperature at any point in time in \algref{get_temp}. The temperature is updated exactly five times a day.

In the following algorithms $ \simutime,$ refers to a timestamp, $\history$ refers to a historical dataset, $K_t$ refers to a set of temperature IDs, and $\weather_{\simutime-1}$ refers to the temperature at the previous time stamp. Here,  $\history=\TrialVone$ and $K_t = \{\text{hot},\text{cold}\}$. The contextual features which influence temperature are time of day, day of the week and the month $tod$, $dow$ and $month$ respectively. Furthermore, at all times besides the first moment in the trial, the next temperature depends on the current temperature $\weather_{\simutime-1}$.

\begin{algorithm}[t!]
	\caption{\textsc{\small{GetTemperature}}\label{get_temp}}
	\begin{algorithmic}[1]
		
		\STATE{$\textsc{INPUT} =  \simutime,\history,K_t,\weather_{\simutime-1},$} 
		 \STATE{$tod\leftarrow tod(\simutime)$}
 		\STATE{$dow\leftarrow dow(\simutime)$}
 		\STATE{$month\leftarrow month(\simutime)$}
		\IF{$\weather_{\simutime-1}$ is Null}
		\STATE{$q \leftarrow [tod,dow,month]$}
		\ELSE
		\STATE{$q \leftarrow [tod,dow,month,\weather_{\simutime-1}]$}
		\ENDIF
		\STATE{$\textbf{p}\leftarrow [0] _{K_l}$}
		\STATE{$\mathcal{T}\leftarrow \statefunctions(\history,q,\weather)$}
		\FOR{$k \in K_t$}
		\STATE{$p_k = \frac{1}{|\mathcal{T}|}\sum_{i=0}^{|\mathcal{T}|}\mathbbm{1}_{l_i==k}$}
		\ENDFOR
		\STATE{$\weather_\simutime \sim Categorical([p_{cold},p_{hot}])$}
		\STATE{\textsc{OUTPUT} $\weather_\simutime$ }	
	\end{algorithmic}
\end{algorithm}

\subsection{Generating location}
In the following algorithms $ \simutime,$ refers to a timestamp, $\groupid_\user$ refers to the group id of user $i$,$\history$ refers to a historical dataset, $K_t$ refers to a set of location IDs, and $\l_{\simutime-1}$ refers to the location at the previous time stamp. Here,  $\history=\TrialVone$ and $K_t = \{\text{other},\text{home or work}\}$. 

As in generating temperature,  
the contextual features which influence location are time of day, day of the week and the month $tod$, $dow$ and $month$ respectively. 
Generating location is different from generating temperature in that each user moves from location to location independently. Whereas we 
model users to share one common temperature, they move from one location to another independently of other users. Thus we also include 
group id in determining the next location for a given user.

\begin{algorithm}[t!]
	\caption{\textsc{\small{GetLocation}}}
	\begin{algorithmic}[1]
		
		\STATE{$\textsc{INPUT} =  \simutime,\groupid_\user,\history,K_l$} 
		 \STATE{$tod\leftarrow tod(\simutime)$}
 		\STATE{$dow\leftarrow dow(\simutime)$}
		\STATE{Find $\starttime$ in $\history$}
		\IF{$\location_{\simutime-1}$ is Null}
		\STATE{$q \leftarrow [tod,dow,\groupid_\user]$}
		\ELSE
		\STATE{$q \leftarrow [tod,dow,\groupid_\user,\location_{\simutime-1}]$}
		\ENDIF
		\STATE{$\mathcal{L}\leftarrow \statefunctions(\history,q,\location)$}
		\STATE{$\textbf{p}\leftarrow [0] _{K_l}$}
		\FOR{$k \in K_l$}
		\STATE{$p_k = \frac{1}{|\mathcal{L}|}\sum_{i=0}^{|\mathcal{L}|}\mathbbm{1}_{l_i==k}$}
		\ENDFOR
		\STATE{$\location_\simutime \sim Categorical([p_{\text{other}},p_{\text{home or work}}])$}
		\STATE{\textsc{OUTPUT} $\location_\simutime$ }
		
	\end{algorithmic}
\end{algorithm}

\subsection{Generating step-counts}
A new step-count is generated for each \patient{} active in the study, every thirty-minutes  according to one of the following scenarios:

\begin{enumerate}
\item \patient{} is at a decision time
\begin{enumerate}
\item \patient{} is available
\item \patient{} is not available
\end{enumerate}
\vspace{-2mm}
\item \patient{} is not at a decision time
\end{enumerate}

Scenarios 1b and 2 are equivalent with respect to how step-counts are generated; a \patient's step count either depends on whether or not they received an intervention (when they are at a decision time and available) or it does not (because they were either not at a decision time or not available).
Recall, that if a user is available the final step count is generated according to \eqnref{steps}.This equation requires sufficient statistics from \TrialVone{}.  The procedure for obtaining these statistics is shown explicitly in \algref{get_steps}.

  \begin{equation}\label{steps}R_{i,k} = \mathbf{N}(\mu_{h(\state_{i,k})},\sigma^2_{h(\state_{i,k})})+ A_{i,k}( f(\state_{i,k})^T\beta_{i} + Z_i). 
  \end{equation}

 \begin{algorithm}[t!]
 \caption{\textsc{StepStatistics} \label{get_steps}}
 \label{alg:cohort}
 \begin{algorithmic}[1]
 \STATE{\textsc{INPUT}  $\text{=} \simutime,\groupid^\user,\weather_\simutime,\user,\history$}
 \STATE{\#Compute variables included in conditioning context}
 \STATE{$tod\leftarrow tod(\simutime)$}
 \STATE{$dow\leftarrow dow(\simutime)$}
 \STATE{$y\leftarrow yst(\simutime,\user)$}
\STATE{$q \leftarrow [\groupid^\user,\weather_\simutime,tod,dow,y,\location_{\simutime,\user},\simaction]$}
\STATE{\#Obtain step counts from $\history$ conditioned on $q$}
\STATE{$\mathcal{S}\leftarrow \statefunctions(\history,q,\stepcount)$}
 \STATE{$\hat{\mu}_{\mathcal{S}} \leftarrow \frac{1}{|\mathcal{S}|}\sum_{i=0}^{|\mathcal{S}|}s_i$}
  \STATE{$\hat{\sigma}^2_{\mathcal{S}} \leftarrow \frac{1}{|\mathcal{S}|}\sum_{i=0}^{|\mathcal{S}|}(s_i-\hat{\mu}_{\mathcal{S}})^2$}
   \STATE{\textsc{OUTPUT}  $\hat{\mu}_{\mathcal{S}},\hat{\sigma}^2_{\mathcal{S}} $}
\end{algorithmic}
 \end{algorithm}
 Here, $\simutime,\groupid^\user,\weather_\simutime,\location_\user,\history$ refer to the current time in the trial, the group id of the $i^{th}$ user, the temperature at time $t$, the location of the $i^{th}$ user, and a historical dataset, respectively. To find sufficient statistics of step counts, we also employ the time of day and day of the week,  $tod$ and $dow$ respectively. 
Finally, $yst(\simutime,\user)$ describes the previous step count as high or low.

\section{Feature construction}
We provide more details on the processes used for feature construction. As stated in the paper we rely heavily on the dataset  \TrialVone{} to make all feature construction decisions. The one exception is in the design of the location feature, for which we had domain knowledge to rely on (more detail below)

\subsection{Baseline activity} Each user is assigned to one of two groups: a low-activity group or a high-activity group. These groups are found from the historical data. We perform hierarchical clustering using the method \textit{hcluster} in  scikit-learn \cite{scikit-learn}. We used a euclidean distance metric to cluster the data and found that two groups naturally arose. These groups were consistent with the population of  \TrialVone{}, which consisted of participants who were generally either office administrators or students. 

\subsection{State features}
We now briefly outline the decisions for the remaining features: time of day, day of the week, and temperature. For each feature we explored various categorical representations. For each, 
the question was how many categories to use to represent the data. For each feature we followed the same procedure. 

\begin{enumerate}
\item We chose a number of categories ($k$) to threshold the data into
\item We partitioned the data into $k$ categories
\item We clustered the step counts according to these $k$ categories
\item We computed the Calinski-Harabasz score of this clustering
\item We chose the final $k$ to be that which provided the highest score 
\end{enumerate}

For example, consider the task of representing temperature. Let $l$ be a temperature, $x$ be a step count and $x_{l_b}$ be a thirty-minute step count occurring when the temperature $l$ was assigned to bucket $b$. Given a historical dataset, we have a vector $\bf{x}$ where each entry $x_{i,t}$ refers to the  thirty-minute step count of user $i$ at time $t$.

\begin{itemize}

\item Let $p$ be a number of buckets. We create $p$ buckets by finding quantiles of $l$. For example, if $p$=2, we find the $50^{th}$ quantile of $l$. A bucket is defined by a tuple of thresholds $(th_1,th_2)$, such that for a data point $d$ to belong to bucket $i$, $d$ must be in the range of the tuple ($th_1\leq d< th_2$).

\item For each temperature, we determine the bucket label which best describes this temperature. That is the label $y$ of $l$, is the bucket for which $th^y_1\leq \bar{s}^l< th^y_2$.

\item We now create a vector of labels $y$, of the same length as $\bf{x}$. Each  $y^l_{i,t}$ is the bucket assigned to $l_{i,t}$. For example, if the temperature for user $i$ at time  $t$ falls into the lowest bucket, $0$ would be the label assigned to $l_{i,t}$. This induces a clustering of step-counts where the label is a temperature bucket. 

\item We determine the Calabrinski-Harabasz score of this clustering. 

\end{itemize}

We test this procedure from $p$ equal to 1, through 4. 

For example, consider determining a representation for time of day. We choose a partition to be morning, afternoon, evening. For each thirty-minute step count, if it occurred in the morning we assign it to the morning cluster, if it occurred in the afternoon we assign it to the afternoon cluster, etc. Now we have three clusters of step counts and we can compute the C score of this clustering. We repeat the process for different partitions of the day. 

\textbf{Time of day} To discover the representation for time of day which best explained the observed step counts, we considered all sequential partitions from length 2-8. We found that early-day, late-day, and night best explained the data. 

\textbf{Day of the week} To discover the representation for day of the week which best explained the observed step counts, we considered two partitions: every day, or weekday/weekend. We found weekday/weekend to be a better fit to the data. 

\textbf{Temperature} Here we choose different percentiles to partition the data. We consider between 2 and 5 partitions (percentiles at 50, to 20,40,60,80). Here we found two partitions to best fit the step counts. We also tried more complicated representations of weather combined with temperature, however for the purpose of this paper we found a simple representation to best allow us to explore the relevant questions in this problem setting. 

\textbf{Location} In representing location we relied on domain knowledge. We found that participants tend to be more responsive when they are either at home or work, than in other places. Thus, we decided to represent location as belonging to one of two categories: home/work or other.

\section{Clinical Trial}
\label{feature_clinical}
In the clinical trial we describe users' states with the features described in Table 4. The two  features which differ from the simulation environment are engagement and exposure to treatment. We clarify these features below. 

\textbf{Engagement}
The engagement variable measures the extent to which a user engages with the mHealth application deployed in the trial. There are several screens within the application that a user can view. Across all users we measure the $40^{th}$ percentile of number of screens viewed on day $d$. If user $i$ views more than this percentile, we set their engagement level to 1, otherwise it is 0.

\textbf{Exposure to treatment}
This variable captures the extent to which a user is treated, or the treatment dosage experienced by this user. Let $D_i$ denote the exposure to treatment  for user $i$. Whenever a message is delivered to a user's phone $D_i$i s updated. That is, if a message is delivered between time $t$ and $t+1$, $D_{t+1}=\lambda D_t+1$. If a message is not delivered, $D_{t+1}=\lambda D_t$. Here, we se $\lambda$ according to data from \TrialVone{} and initialize $D$ to 0.

\bibliography{pooling}
\bibliographystyle{icml2020}